# Optimal Factory Scheduling using Stochastic Dominance A*

Peter R. Wurman and Michael P. Wellman
University of Michigan
Artificial Intelligence Laboratory
1101 Beal Avenue
Ann Arbor, MI, 48109-2110
{pwurman, wellman}@umich.edu

## Abstract

We examine a standard factory scheduling problem with stochastic processing and setup times, minimizing the expectation of the weighted number of tardy jobs. Because the costs of operators in the schedule are stochastic and sequence dependent, standard dynamic programming algorithms such as A* may fail to find the optimal schedule. The SDA* (Stochastic Dominance A*) algorithm remedies this difficulty by relaxing the pruning condition. We present an improved state-space search formulation for these problems and discuss the conditions under which stochastic scheduling problems can be solved optimally using SDA*. In empirical testing on randomly generated problems, we found that in 70%, the expected cost of the optimal stochastic solution is lower than that of the solution derived using a deterministic approximation, with comparable search effort.

## 1 INTRODUCTION

Generating production schedules for manufacturing facilities is a problem of great theoretical and practical importance. The Operations Research and Artificial Intelligence communities have studied various versions of this problem. During the last decade, an effort has been made to understand the relationships between the techniques developed by these two fields. The work described here aims to continue in this vein by showing how a class of well-defined scheduling problems can be mapped into a general search procedure.

In particular, we are concerned with generating static schedules over a limited horizon in a multi-product factory with a single, bottleneck machine whose performance is specified by stochastic processing times and sequence-independent, stochastic setup times. We refer to this model as the stochastic lot-sizing problem. The demand on the factory is specified by a set of orders for products with deadlines and tardy penalties. The challenging scheduling problems occur when demand is greater than capacity.

Although uncertainty in processing times is widely acknowledged, few approaches produce strictly optimal schedules in stochastic lot-sizing problems. We present a general approach based on SDA* (Stochastic Dominance A*), a state-space search algorithm designed for uncertain, path-dependent costs (Wellman, Ford, and Larson 1995). To apply SDA* to scheduling problems, we must extend the algorithm to handle multidimensional cost structures.

In the next section we provide motivation for the problem. Section 3 reviews SDA*, and Section 4 describes the formulation of the factory scheduling problem in detail, including the multidimensional extensions to SDA*. In Section 5 we discuss the details of our implementation of the problem, including the heuristics used and the empirical results. The sixth and seventh sections discuss related work and possible future directions of this research, respectively.

## 2 PRODUCTION SCHEDULING UNDER UNCERTAINTY

Consider a factory with one machine that is capable of making three products: X, Y, and Z, with mean processing times given in Table 1 (in minutes). The machine is currently set up to build product X. Initially, the factory has two orders due at the end of day 1 (minute 480), as shown in Table 2. If an order is not shipped by the time it is due, it incurs a late penalty. The late penalties of order R1 and R2 are $w_1$ and $w_2$, respectively.

At the start of the day the factory needs to construct a sequence in which to process the jobs. We assume

Table 1: Production Data

| Product | Run time | Setup time |
|---------|----------|------------|
| X | 20 | 5 |
| Y | 30 | 10 |
| Z | 25 | 5 |

Table 2: Order Information

| Order | Product | Quantity | Deadline | Penalty |
|-------|---------|----------|----------|---------|
| R1 | X | 11 | 480 | $w_1$ |
| R2 | Y | 8 | 480 | $w_2$ |

that the schedule created is static and jobs cannot be preempted from the machine. For this simple problem, there are two possible sequences:

$A_1$  Build R1→Setup Y→Build R2
$A_2$  Setup Y→Build R2→Setup X→Build R1

If the mean times specified above are taken as certain, then it does not matter which sequence is followed. $A_1$ takes 470 minutes, $A_2$ 475 minutes—both finish by the end of the day.

However, if there is uncertainty about the run and setup times, then it may matter which schedule we use. Given the possibility that we will not complete both orders on time, we must consider both the probability of being late and the potential late penalty associated with each sequence. For example, if the run times of the products are normally distributed with means as above and standard deviations of 2 minutes, then $A_1$ has a 0.125 probability of being late, while $A_2$ has a 0.284 probability of being late. Which schedule is optimal depends on the relative penalties of being late on R1 versus R2. In this case, $A_2$ has a lower expected penalty than $A_1$ iff $w_1 < 0.44w_2$.

Now consider the case where we have a third order, R3, for 19 units of product Z due at the end of day 2 (960 minutes). Let $A_1'$ and $A_2'$ denote the schedules that extend $A_1$ and $A_2$, respectively, by appending a 'Setup Z' followed by a 'Build R3'. The probability of shipping R3 late is 0.209 for $A_1'$ and 0.322 for $A_2'$.

Clearly, if $A_1$ has a lower expected penalty than $A_2$ with respect to the first two orders, it is going to continue to be better when both are extended to build R3, since it has a lower probability of being late on that order. However, if $A_1$ has a higher expected penalty than $A_2$, we must analyze the complete schedules to determine which performs better with the extension. (In fact, if the penalty associated with R3 is high enough, the optimal sequence may be neither of these.)

If states are defined by the jobs we have processed, we cannot generally construct a schedule incrementally by extending the best partial solutions. Therefore, the straightforward application of dynamic programming, in which only the best path to each intermediate state is retained, would not be valid. We could recover validity by including the time at which a sequence completes as part of the state, but this would dramatically increase the state space, reducing the effectiveness of dynamic programming in defeating the combinatorics.

This example illustrates the two principal issues we address in this work. First, scheduling using a deterministic approximation based on the means of random variables can produce suboptimal solutions. Second, solutions that have the earliest expected completion time do not necessarily have the lowest expected penalty. Straightforward approaches to retain optimality by encoding additional features in the state can lead to an unacceptable explosion of the state space. The SDA* algorithm performs optimal scheduling using the stochastic model directly, and our mapping of scheduling problems into the SDA* framework maintains time information outside the state encoding.

## 3  STOCHASTIC DOMINANCE A*

A* is a well-understood state-space search technique that guarantees optimal paths when the operator costs are deterministic and solely a function of the current state. However, when costs are stochastic and path dependent, A* may prune partial paths that could lead to superior solutions.

### 3.1  PATH DEPENDENCE

Path-dependent costs occur in situations where the cost of applying operator $a$ in state $S$ depends upon the cost of the path taken to reach that state. The *operator cost* function is given by $c(a, S, C)$, where $C$ is the cost of the path taken to $S$.[1] One source of path dependence is a utility function that is nonlinear in time, such as a binary utility function based on meeting a deadline. Formally, let $A = <a_1, \ldots, a_k>$ be a sequence of actions, and let $a_j(S)$ denote the state resulting from applying action $a_j$ in state $S$. If $S_0$ is the initial state, then executing sequence (or path) $A$ results in state

$$A(S_0) = a_k(a_{k-1}(\ldots(a_1(S_0))\ldots)).$$

The *path cost*, $C(A, S_0)$ of executing $A$ from state $S_0$ can be expressed recursively. Let $A^j$ be the sequence

---
[1] We assume for the nonce that costs are represented by scalar quantities, totally ordered by $\leq$. In addition, we assume throughout the paper that utility is nonincreasing in cost.

defined by the first $j$ actions of $A$, and $S_j = A^j(S_0)$,

$$\mathcal{C}(A^j, S_0) = \mathcal{C}(A^{j-1}, S_0) + c(a_j, S_{j-1}, \mathcal{C}(A^{j-1}, S_0)).$$

(Henceforth we omit the state argument when the initial state is unambiguous.)

It has been shown (Kaufman and Smith 1993) that A* produces optimal solutions even with path-dependent cost functions, as long as a particular *consistency*, or *monotonicity*, condition applies. The monotonicity condition demands that for any path costs $\mathcal{C} \leq \mathcal{C}'$,

$$\mathcal{C} + c(a, S, \mathcal{C}) \leq \mathcal{C}' + c(a, S, \mathcal{C}'). \quad (1)$$

Note that this form of monotonicity—on the accumulated path cost—is weaker than requiring that $c$ be monotonic in $\mathcal{C}$.

It follows from this condition that for two paths, $A$ and $A'$, leading to the same state, the superiority of one, $\mathcal{C}(A) \leq \mathcal{C}(A')$, implies that the same relation holds for these paths extended by a given action, $a$. That is, $\mathcal{C}(Aa) \leq \mathcal{C}(A'a)$. Given this result, it is safe to prune $A'$ because, for any path to the goal based on that path, there is a path at least as good based on $A$.

### 3.2 STOCHASTIC COSTS

A second variation of standard state-space search is to admit stochastic costs, that is, to treat $c$ as a random variable. If $c$ depends only on the state, and utility is linear in cost (i.e., the agent is risk neutral), then it is sufficient to use A* with operator costs represented by their means.

However, if the problem requires both stochastic and path-dependent operator costs, then we are no longer justified in pruning paths based upon expected costs. In such cases we can use the Stochastic Dominance A* (SDA*) algorithm. SDA* is a variation of A* with the following four enhancements.

**Stochastic Monotonicity:** We require a stochastic version of the monotonicity condition used to address path dependence in the deterministic case. *Stochastic dominance*, indicated by $\leq_{SD}$, is the appropriate comparator. A random variable $x_1$ stochastically dominates another random variable, $x_2$, if, for all $z$,

$$\Pr(x_1 \leq z) \geq \Pr(x_2 \leq z). \quad (2)$$

From (1) and (2) we define the stochastic monotonicity condition. For all costs $\mathcal{C}$, $\mathcal{C}'$, and $z$, $\mathcal{C} \leq_{SD} \mathcal{C}'$,

$$\Pr(\mathcal{C} + c(a, S, \mathcal{C}) \leq z) \geq \Pr(\mathcal{C}' + c(a, S, \mathcal{C}') \leq z). \quad (3)$$

**Pruning:** Rather than keeping the single lowest-cost path to a node, we must keep all of the *admissible* paths, where admissibility is defined by stochastic dominance. We have previously shown (Wellman, Ford, and Larson 1995) that if paths $A$ and $A'$ lead to the same state and $\mathcal{C}(A) \leq_{SD} \mathcal{C}(A')$, then $A'$ cannot be part of a uniquely optimal solution. Specifically, the stochastic monotonicity condition (3) in this situation entails that for any incremental action $a$, $\mathcal{C}(Aa) \leq_{SD} \mathcal{C}(A'a)$.

If, however, $A'$ is not stochastically dominated, then it is possible to construct an example where it does in fact lead to the optimal solution.

**Heuristics:** Whereas a heuristic is admissible for A* if it underestimates the cost to the goal, for the stochastic path-dependent case an admissible heuristic must produce estimated cost distributions that stochastically dominate the actual cost distribution. In addition, the heuristics can be functions of the path cost as well as the state.

**Priority:** Search nodes are expanded in order of estimated expected utility. Like A*, the algorithm terminates when a goal node is popped off the priority queue. The reasoning is as follows: given that the heuristic function is stochastically admissible, and the accumulated path costs stochastically monotone, expected utility is monotonically decreasing along a path. Thus, when a solution is found, any intermediate path that had an estimated expected utility less than that of the solution must have already been explored or pruned.

Under the conditions described above, SDA* provides an optimal and complete solution procedure for problems with path-dependent stochastic operator costs. The relation of SDA* to these problem features is summarized in Table 3. In this table, we see that path-dependence alone can be accommodated by a monotonicity condition, and stochastic costs alone by using means, but the conjunction of both requires SDA*.

Table 3: Appropriate Search Methods for Scalar Costs

|  | State Dependent | Path Dependent[2] |
|---|---|---|
| Deterministic Costs | A* | A* |
| Stochastic Costs | A* with means | SDA* |

In the factory scheduling problem, we wish to avoid encoding time in the state. To accomplish this, we define states by the jobs completed and use a two-dimensional cost structure that captures both the time and penalty. The extension of A* to multidimensional costs has already been investigated by Stewart and White (1991), who proposed the Multiobjective A* (MOA*) algorithm for this case. Like SDA*, MOA* extends A* by pruning paths based on dominance rather than point utility. This technique can be extended to stochastic and path-dependent costs in a manner analogous to the scalar case, as we diagram in Table 4. The particular contribution of this paper lies in the lower right cell of this table.

---

[2] All path-dependent cases for Tables 3 and 4 require a (deterministic or stochastic) monotonicity condition.

Table 4: Appropriate Search Methods for Multidimensional Costs

|                     | State Dependent | Path Dependent[2] |
|---------------------|-----------------|-------------------|
| Deterministic Costs | MOA*            | MOA*              |
| Stochastic Costs    | MOA* with means | MO-SDA*           |

## 4  PROBLEM FORMULATION

In this section we discuss the details of the multidimensional extension of SDA* for finding the optimal solution in the stochastic scheduling problem.

### 4.1  NOTATION

Consider a factory with $n$ orders for $m$ products. Each product has one probability distribution that defines its processing time and another that defines the amount of time it takes to set up the machine. For descriptive simplicity, setups are assumed to be sequence independent, that is, the amount of time it takes to change the machine setup from product $i$ to product $j$ is independent of the the value of $i$, for $i \neq j$.

Each order is defined by the tuple ⟨*product, quantity, deadline, penalty*⟩. We will use the following additional notation:

$X_{iq}$ = stochastic processing time to make $q$ units of product $i$
$\Delta_i$ = stochastic setup time of product $i$
$b_j$ = product of order $j$
$q_j$ = quantity of order $j$
$d_j$ = deadline of order $j$
$w_j$ = penalty for shipping order $j$ late
$b(S)$ = the setup of the machine in state $S$
$r_i(S)$ = inventory of product $i$ at state $S$
$o_j(S)$ = status of order $j$ in state $S$ (either shipped or unshipped)
$T(A)$ = time distribution of path $A$
$W(A)$ = accrued stochastic penalty of path $A$

### 4.2  STATES

We encode a state as a combination of the current inventory, the machine setup, and the status of the orders. The inventory is a list of the quantity of each product that has been produced but not shipped. The status of the orders is a list that indicates whether each order has been shipped.

The initial state specifies all unshipped orders, some initial inventory, and an initial machine setup. A solution is any state in which all of the orders are shipped.

### 4.3  OPERATORS

There are three types of operators that the factory can execute: make products, ship orders, or change the machine setup. We define each operator by properties of the state $S'$ resulting from applying the operator in state $S$.

The **make** operation converts raw materials into finished inventory. The only product that can be made is the one for which the machine is set up. Make $q$ units of product $i$, where $i = b(S)$, results in inventory

$$r_i(S') = r_i(S) + q.$$

The **setup** operator for product $i$ has the effect

$$b(S') = i.$$

Finally, **ship** order $j$ removes the corresponding amount of inventory and packages it up for a customer:

$$o_j(S') = shipped$$
$$r_i(S') = r_i(S) - q_j$$

We discuss methods for restricting the states in which these operators are applicable in Section 5.1.

### 4.4  COSTS

In the path-planning problem presented by Wellman et al. (1995), utility is inversely related to time, which means the path with the lowest expected time to the goal has the highest utility. In the lot-sizing problem our objective is minimizing expected penalty. Although the penalty is a function of the time each order is completed, the path with the smallest expected time does not necessarily have the lowest expected penalty.

We specify the cost of a path in the lot-sizing problem by the pair ⟨*penalty, time*⟩, that is, $C(A) \equiv \langle W(A), T(A) \rangle$. The following equations define the cost effects of the three operators when extending the path from $A$ to $A'$.

The make operator incurs no penalty but does have an effect upon the current time. To make $q$ units of product $i$ costs:

$$T(A') = X_{iq} \oplus T(A),$$

where $X_{iq} \oplus T(A)$ is the distribution corresponding to the sum of random variables $X_{iq}$ and $T(A)$.

The setup operator also has only a time effect. To set up the machine to build product $i$:

$$T(A') = \Delta_i \oplus T(A).$$

If the setup operation were sequence dependent, we would simply replace $\Delta_i$ with $\Delta_{hi}$ in the above equation, where $\Delta_{hi}$ is the time necessary to change the machine from product $h$ to product $i$.

The ship operator has no effect on time, affecting only the penalty component of cost. The incremental penalty for shipping an order is $w_j$ if the shipping time is past the deadline (i.e., the order is late), and zero otherwise. The calculation of the accrued penalty distribution is complicated by the fact that the incremental penalty is not independent of previous penalties since they are all derived from the underlying time distribution. However, when utility is linear in total penalty, we can simply keep track of *expected* penalty. The expected value of the accrued penalty after shipping order $j$ is given by:

$$E[W(A')] = E[W(A)] + w_j \Pr(T(A) > d_j). \quad (4)$$

For the make and setup operators, the incremental costs are path-independent. For the ship action, the cost increases monotonically as a function of time. With these properties, the stochastic monotonicity condition (3) is met.

Our two-component cost measure puts us in the realm of multiobjective search. In the MOA* algorithm described by Stewart and White (1991), paths to a state can be pruned only when their costs are dominated by an existing path to that state. Cost vector $V$ dominates vector $V'$ iff each element of $V$ is less than or equal to the corresponding element of $V'$, and at least one element of $V$ is strictly less than the corresponding element in $V'$. Our case is somewhat more complicated, as one of the elements is a random variable, and the second element is path-dependent on the first. Specifically, the penalty element is a function of the (stochastic) time element. Therefore we must prove for this situation that the multidimensional extension to SDA* prunes only nodes that cannot lead to a lower-cost solution. Although our theorem is stated in terms of our particular factory scheduling problem, the result holds for the more general case of multidimensional, stochastic path-dependent costs, given the stochastic monotonicity condition.

**Theorem 1** *Let $A$ and $A'$ be two paths to state $S$. Path $A'$ can be safely pruned iff $E[W(A)] \leq E[W(A')]$ and $T(A) \leq_{SD} T(A')$.*

*Proof.* (If) Consider two paths satisfying the theorem's condition. We consider extensions to these paths formed by adding operator $a$.

First let $a = \text{make}(i, q)$. In this case, $T(Aa) = X_{iq} \oplus T(A)$, and $T(A'a) = X_{iq} \oplus T(A')$. Similarly, if $a = \text{setup}(i)$, then $T(Aa) = \Delta_i \oplus T(A)$, and $T(A'a) = \Delta_i \oplus T(A')$. It is straightforward to show that for any probability distributions $f$, $f'$, and $g$,

$$f \leq_{SD} f' \Leftrightarrow g \oplus f \leq_{SD} g \oplus f'. \quad (5)$$

Therefore, for both make and setup operators, stochastic dominance of the time distributions is preserved. Neither operator type affects accrued penalty.

Next, suppose $a = \text{ship}(j)$. This operator has no effect on time, but does modify expected penalty. Specifically, appending $a$ to our two paths yields $E[W(Aa)] = E[W(A)] + w_j \Pr(T(A) > d_j)$, and $E[W(A'a)] = E[W(A')] + w_j \Pr(T(A') > d_j)$. By the definition of stochastic dominance (2), for any $d_j$,

$$T(A) \leq_{SD} T(A') \Rightarrow$$
$$\Pr(T(A) \geq d_j) \leq \Pr(T(A') \geq d_j).$$

Since $E[W(A)] \leq E[W(A')]$, this gives us $E[W(Aa)] \leq E[W(A'a)]$.

We have therefore shown that for any operator, adding the operator to our two paths preserves the inequality on expected penalty and the stochastic dominance of the time element of cost. By induction, this will remain true for any sequence of operators, and thus any extension of $A'$ has a corresponding dominating extension in $A$. Therefore, pruning $A'$ cannot eliminate a uniquely optimal solution.

(Only If) There are two cases for which $A$ does not dominate $A'$:

1. $E[W(A)] > E[W(A')]$
2. $T(A) \not\leq_{SD} T(A')$

We demonstrate that pruning in either of these cases can lead to suboptimal solutions by showing that for any non-goal state, there is some setting of deadlines and penalty weights such that $A'$ has an extension with greater expected utility than any extension of $A$.

Suppose the first case, and let $\alpha$ be a sequence of actions that leads to the goal from state $S$. Whatever the respective time distributions, $T(A)$ and $T(A')$, it is possible to set the deadlines and penalty weights of $S$'s unfilled orders such that the remaining penalty accrued by $\alpha$ is the same whether appended to $A$ or $A'$. For example, set the deadlines so that they are already past, and all remaining orders are necessarily late. Or set the deadlines sufficiently far away, or penalty weights sufficiently low, so that the remaining penalty is negligible. In either case, the fact that $A'$ has lower penalty than $A$ entails that $A'\alpha$ also has a lower expected penalty than $A\alpha$.

Next consider the second case, where $A$ does not stochastically dominate $A'$ in time. Consider an order, $j$, for product $i$, that is unfilled in state $S$. Let $q' = q_j - r_i(S)$ and $\alpha$ be the minimal sequence of actions necessary to reach a state, $S'$, in which $j$ can be shipped. If $b(S) = i$ then we need to perform a **make** action and $T(A\alpha) = T(A) \oplus X_{iq'}$. If $b(S) \neq i$, then we must perform a **setup** action as well, and $T(A\alpha) = T(A) \oplus X_{iq'} \oplus \Delta_i$. In either case, we can conclude from (5) that

$$T(A) \not\leq_{SD} T(A') \Rightarrow T(A\alpha) \not\leq_{SD} T(A'\alpha).$$

By the definition of stochastic dominance,

$$T(A\alpha) \not\leq_{SD} T(A'\alpha) \Rightarrow$$
$$\exists t.\; \Pr(T(A\alpha) \geq t) > \Pr(T(A'\alpha) \geq t). \quad (6)$$

Let $d_j$ equal a value of $t$ for which (6) holds, and let $p = \Pr(T(A\alpha) \geq d_j) - \Pr(T(A'\alpha) \geq d_j) > 0$. The difference in penalty for the extended paths is then

$$E[W(A\alpha)] - E[W(A'\alpha)] =$$
$$E[W(A)] - E[W(A')] + pw_j. \quad (7)$$

Let $w_j$ take on a value for which $pw_j > \Sigma_{i \neq j} w_i$. The right hand side of (7) is then positive, which implies $E[W(A\alpha)] > E[W(A'\alpha)]$. Since with this large setting of $w_j$ the extension by $\alpha$ is necessarily optimal (i.e., by making order $j$ far more important than the rest, the best policy is to produce it next), we have a case where pruning $A'$ can eliminate potentially optimal solutions. This concludes the proof. □

To ensure that the SDA* algorithm will produce optimal solutions with our two-dimensional cost structure, we must also show that the priority function and the termination condition are valid.

Since the utility is linear solely in the penalty, the priority function expands nodes in increasing order of their estimated expected penalty. This ensures that when a goal node is popped from the queue, all partial solutions remaining have an expected penalty at least as great. Since the heuristic evaluation is guaranteed to be an underestimate, and no operator decreases penalty, the first goal node popped off the queue must be an optimal solution.

## 5 EMPIRICAL STUDIES

### 5.1 OPERATOR APPLICABILITY

The three operators defined above are really operator schema. A crucial step in the design process is defining how operators are instantiated in each state so as to restrict the state space to feasible and non-trivial variations of the production sequence. One possible method for generating operators is to define a set of make operators of fixed quantities. This would model a manufacturing environment in which batch sizes are fixed. Alternately, we could have a joint **make-&-ship** operator for each order. While this mechanism allows us to do away with the **ship** operator as a separate step, it makes the number of operators proportional to the number of orders, $n$. We present a method designed to keep the branching factor linear in the number of products, $m$, without reducing the descriptiveness of the search space.

**Setup** actions are allowed only when the total inventory is zero. They can change the setup only to products with unmet demand. We also prevent sequences with two **setup** actions in a row. This restricts **setup** operators to the initial state and immediately after an order is shipped.

**Ship** operations are allowed only if there is exactly the correct amount of product in inventory to meet the needs of an order.

From each state only one **make** action is allowed: we limit **make** operators to produce only the minimum quantity of a product that will allow a new **ship** action. In other words, if we have $r_i$ units of product $i$ in inventory in state $S$, then the **make** operator that will be generated is to produce quantity, $q*$, such that

$$q* = \min_J (q_j - r_i(S)),$$
$$\text{where } J = \{j | b_j = i, o_j(S) = unshipped,$$
$$\text{and } q_j > r_i(S)\}.$$

For example, consider the set of orders given in Table 5. Assume that we start at state $S_0$, with no inventory and with the machine set up to produce product X. The available operators are simply '**Make-4-of-X**' and '**Setup-Y**'.

Figure 1 shows the operators allowed in the initial region of the state space for these orders. Notice that in state $S_2$ we are restricted to shipping order R4 even though we have more than enough to ship R5. We made the decision about whether the first four units were destined for order R4 or R5 in state $S_1$.

Table 5: Sample Orders

| Order | Product | Quantity | Deadline |
|-------|---------|----------|----------|
| R4 | X | 6 | 480 |
| R5 | X | 4 | 480 |
| R6 | Y | 2 | 480 |
| R7 | Y | 6 | 480 |

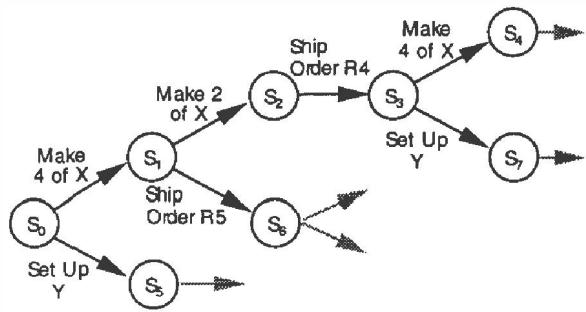

Figure 1: State Space Showing the Restricted Application of Operators

Imposing these restrictions entails no loss of generality. Delaying a ship operator can never result in a reduced final penalty, thus it is never to our advantage to produce more than the amount of the order before shipping the order. Likewise, it is never advantageous to change the machine setup before shipping an order.

The operator rules that we have chosen are designed to bound the branching by $m + 1$, rather than the (potentially much larger) $n$.[3] It allows the algorithm to examine the efficacy of producing any amount of a product without committing that product to a specific order. This reduction in branching factor comes at the expense of increased solution depth. We have not performed any empirical studies to compare the efficiency of these operator rules against, say, the `make-&-ship` rules. The guarantee of optimality is not affected by the choice of operator schemes.

## 5.2 HEURISTICS

One of the benefits of casting the stochastic scheduling problem as a form of A* search is that we can use our experience in developing heuristics for the deterministic single-machine scheduling domain. We look for heuristics that are guaranteed underestimates of, and significantly easier to compute than the actual remaining path cost. A useful technique for finding heuristics is to examine relaxations of the problem. Using this method, we have implemented two heuristics for this scheduling domain.

---

[3]Strictly speaking, if there are multiple orders for exactly the same quantity of the same product, the branching factor can exceed this bound by the number of orders in such an equivalent set.

### 5.2.1 Parallel-Machines Heuristic

The first heuristic, which we call the Parallel-Machines heuristic, relaxes the assumption that orders must be processed in series. This heuristic estimates the remaining penalty by summing the incremental penalty that each order would incur if it shipped next. Since only one order can actually be next, the rest will necessarily ship at a later time and incur at least as much incremental penalty. Therefore, the heuristic is guaranteed not to overestimate the actual remaining penalty. The advantage of this heuristic is that it can apply the full penalty cost for late orders and account for setup times. The disadvantage is, of course, that it ignores the impact that processing one order will have on the other orders.

### 5.2.2 Fractional-Penalty Heuristic

The second heuristic, which we call the Fractional-Penalty heuristic, relaxes the assumption that orders must be shipped in full. By assuming that fractional orders can be shipped, and therefore a portion of the late penalty avoided, we can estimate the penalty of all of the remaining orders processed serially. To make the heuristic computationally efficient, we also relax the machine setup requirements and limit the horizon of the estimate to a single deadline. Looking farther than a single deadline or accounting for setup times would require examining many combinations of actions to guarantee an underestimate of the overall cost.

The lot-sizing problem without setup requirements is equivalent to the job scheduling problem, the deterministic version of which is analogous to the 0-1 KNAPSACK problem. Kolesar (1967) presented the FRACTIONAL KNAPSACK problem as an effective search heuristic for the 0-1 KNAPSACK problem. This insight gives us a strategy for computing the fractional-penalty heuristic with a greedy algorithm. Like the algorithm for the FRACTIONAL KNAPSACK problem, the fractional-penalty heuristic sorts the orders by a measure of their value: the penalty per unit time. Under the assumption that orders are divisible and setup times are zero, processing the remaining jobs in decreasing order of their penalty per unit time is guaranteed to underestimate the sum of the remaining penalty.

Table 6: Nova Production Environment

| Product | Run Time | | Setup Time | |
|---|---|---|---|---|
| | $\mu$ | $\sigma$ | $\mu$ | $\sigma$ |
| 1 | 2.9 | .2 | 5 | .1 |
| 2 | 3.1 | .2 | 5 | .1 |
| 3 | 3.1 | .2 | 12 | .1 |
| 4 | 3.4 | .3 | 15 | .2 |
| 5 | 3.7 | .3 | 15 | .2 |
| 6 | 4.0 | .4 | 15 | .2 |

### 5.3 EMPIRICAL RESULTS

The data with which we tested our algorithm is adapted from a simulation of a hypothetical corporation called Nova, Inc. (Muckstadt and Severance 1995). Nova's factory manufactures six products with the production data summarized in Table 6. We considered the processing and setup times to be normal distributions truncated at ±4 standard deviations. This allowed efficient convolution and stochastic dominance calculations.

Our empirical investigation focused on comparing the stochastic version of the problem, solved using the SDA* algorithm, to the deterministic model. We generated 700 random problems with between 6 and 20 orders each and total estimated capacities of between 95 and 125 percent (including setup times). Setup actions from the initial state were not allowed.

To compare the solutions found by both models, we solved the same problem using both stochastic and deterministic models of the data. We then applied the schedule found by the deterministic model to the stochastic data and compared its expected cost to that of the optimal schedule produced by the stochastic model. The stochastic schedule is always at least as good as the deterministic schedule, but we found that in approximately 72% of the over-constrained problems (total capacity > 100%) the stochastic solution was strictly better. On average, the expected penalty was reduced by 15%. The penalty as a function of capacity is shown in Figure 2. We expect that frequency and magnitude of the improvements are a function of the penalty function and the variance of the processing and setup times.

To get a feel for the behavior of the SDA* formulation versus the deterministic formulation, we compared the number of nodes expanded and the number of nodes pruned for the 700 problems described above. We found both measures to be very similar to the deterministic model. The exception to this observation is, predictably, the region around 100% capacity. Since the random problems are generated to approximate a certain capacity, many problems in the 100% region will be require less than the full capacity and will be easy deterministic problems. However, approaches that take into account the stochastic nature of the processes will be sensitive to penalties in problems where the mean capacity is slightly below 100%. Figure 3 shows graphs of expansions and dominations for 20 order problems.

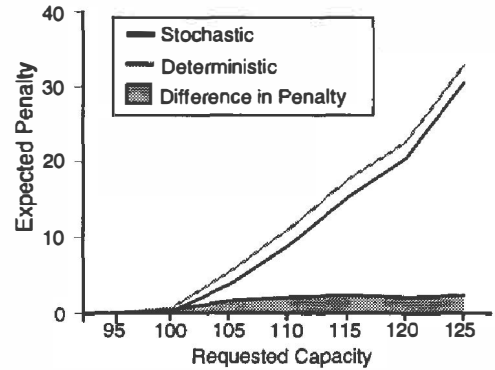

Figure 2: Expected Penalty as a Function of Capacity

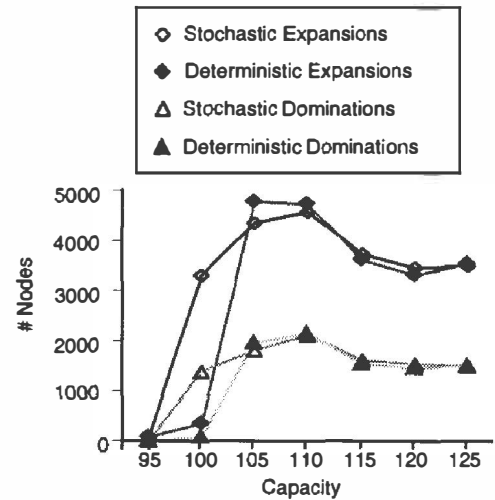

Figure 3: Number of Expansions and Dominations at Varying Estimated Capacities for Randomly Generated 20-Order Problems

The improved performance for a fixed number of orders as the capacity increases past 110% can be explained as improved estimates by the heuristics as the average order size increases.

## 6 RELATED WORK

Recent years have seen an increase in cooperation between the fields of Operations Research and Artificial Intelligence, especially in the area of scheduling (Zweben and Fox 1994; Brown and Scherer 1995). For comprehensive surveys of the major scheduling approaches in both fields, see (Brown, Marin, and Scherer 1995) and (White 1990).

Most work in OR is focused on the job scheduling problem: a subclass of the lot-sizing problem in which the setup time can be lumped with the processing time. With this simplification, the time necessary to execute a schedule is completely specified by the list of jobs completed and is independent of the sequence in which the jobs are executed.

Deterministic, single-machine problems have received a great deal of attention in the literature. The computational complexity varies greatly with the problem assumptions. For example, the problem of minimizing the total weighted completion time in job scheduling, has, as an optimal policy, the weighted shortest processing time (WSPT) rule. However, the task of minimizing the total weighted tardiness for the same problem is well known to be an instance of the 0-1 KNAPSACK problem (Nuttle and Aly 1986). Schrage (1981) classifies variations of the deterministic lot sizing problem, and shows that many of them are NP-hard.

The complexity hierarchy for stochastic problems is less fully characterized than that for deterministic problems. In many instances, the stochastic versions of scheduling problems are harder than the deterministic ones. For selected objective functions, the optimal policies for deterministic job scheduling problems are optimal for their stochastic counterparts (Crabill and Maxwell 1969). However, under the assumption that processing times are exponentially distributed, some problems whose deterministic versions are NP-complete have been shown to have polynomial time stochastic solutions (Pinedo 1981; Derman, Lieberman, and Ross 1978).

The search algorithm used in this paper is closely related to a variety of recent work in multicriteria search. Carraway, et al..(1990) proposed *generalized dynamic programming* as a method for finding optimal solutions to problems where utility is a function of multiple deterministic attributes. This concept was extend to A* search by adding heuristics (White, Stewart, and Carraway 1992). Stewart and White (1991) proposed multiobjective A* (MOA*) for cases where the attributes cannot be mapped into a single utility function, though it works equally well for cases where the utility can be fully evaluated in goal states but not in intermediate states. Loui (1983) noted that some stochastic problems reduce to deterministic multiobjective problems when the distributions are uniquely determined by a set of parameters from which dominance can be established. Wellman, et al. (1995) established stochastic dominance as an appropriate pruning condition for the PFS-Dominance and SDA* algorithms. This paper builds upon that work by extending SDA* to the multidimensional case.

All multicriteria formulations relax dynamic programming to allow a set of undominated costs at each node in the network. This relaxation opens the door for operator costs that vary depending upon the path taken to a node. The need for a consistency condition to retain optimality when costs are path dependent was recognized by Kaufman and Smith (1993), and extended to the stochastic case in (Wellman, Ford, and Larson 1995).

# 7  FUTURE WORK

Although the analysis above focuses on one specific class of scheduling problems, the technique described is more widely applicable. It could be applied with little modification to variations of the lot sizing problem such as those that have sequence-dependent setup times, or problems that have setup penalties (perhaps related to non-reusable tooling costs).

Other interesting variations could be formulated in a similar way, but would require clever extensions to the operator generation procedure in order to keep the branching factor down. One such class of problems are models with less than perfect yield rates. Tardy penalties would be functions of both time and yield. The optimization question becomes one of which products should be overproduced and by how much in order to minimize the expected tardiness penalty. Problems with nonzero release times could be addressed by taking the distribution that represents the upper bound between the release time and time element of the path cost. The multiple machine problem could be addressed by encoding multiple time distributions in the path cost. Generating operators to distribute the work load between machines would be especially challenging.

Although the algorithm is easily adapted for any penalty function that is nondecreasing in time, we have thus far examined its behavior only with the weighted number of tardy jobs. This was done to most easily compare it against other techniques. We have not systematically studied the effect of varying the number of products, or using penalties other than the number of units in the order. Thus, although our experience provides some intuition about the behavior of the algorithm, we cannot yet draw any wide-ranging conclusions.

## 8 CONCLUSION

We have shown that scheduling problems with objective functions that are nondecreasing and nonlinear in time can be formulated using a multidimensional cost structure. This formulation produces costs that are path-dependent, mandating that a monotonicity condition hold for pruning to be valid. When the operators have stochastic effects, we can use a modified version of SDA* to find the optimal solution. As an example, we presented a formulation for the problem of generating an optimal static schedule for a multi-product, single-machine environment with tardy penalties. Our empirical investigation showed that in a significant number of sample problems, schedules could be improved by accounting for stochastic effects.

Most importantly, the formulation we have presented is applicable to a wide variety of stochastic problems that are often overlooked in the literature because they have, until now, been difficult to formulate as state-space search. Our method finds optimal solutions to the difficult stochastic lot-sizing problem with less restrictive assumptions on probability distributions than have been previously considered, requiring only the relatively benign assumption of stochastic monotonicity.

### Acknowledgements

We are grateful to the anonymous reviewers and students in the U-M Decision Machines Group for suggestions about the presentation of this work. This work was supported in part by Grant F49620-94-1-0027 from the Air Force Office of Scientific Research. Additional support came from the Horace H. Rackham School of Graduate Studies through a research partnership with Prof. Dennis Severance.